\pdfoutput=1 
\documentclass[letterpaper, 10 pt, conference]{ieeeconf}  
\usepackage{graphicx}
\usepackage{multirow}
\usepackage{amsmath}
\usepackage{amsfonts}
\usepackage{makecell}

\IEEEoverridecommandlockouts                              

\title{\LARGE \bf
A Semi-Supervised Classification Method of Apicomplexan Parasites \\and Host Cell Using Contrastive Learning Strategy
}

\author{Yanni Ren*, Hangyu Deng*, Hao Jiang* and Jinglu Hu*
\\
\{yanni, deng.hangyu, haojiang\}@fuji.waseda.jp,  jinglu@waseda.jp
\thanks{*The authors are with the Graduate School of Information, Production and Systems, Waseda University, Kitakyushu-shi, 808-0135 Japan.}
}

\UseRawInputEncoding
\begin{document}

\maketitle
\thispagestyle{empty}
\pagestyle{empty}

\begin{abstract}

A common shortfall of supervised learning for medical imaging is the greedy need for human annotations, which is often expensive and time-consuming to obtain. This paper proposes a semi-supervised classification method for three kinds of apicomplexan parasites and non-infected host cells microscopic images, which uses a small number of labeled data and a large number of unlabeled data for training. There are two challenges in microscopic image recognition. The first is that salient structures of the microscopic images are more fuzzy and intricate than natural images' on a real-world scale. The second is that insignificant textures, like background staining, lightness, and contrast level, vary a lot in samples from different clinical scenarios. To address these challenges, we aim to learn a distinguishable and appearance-invariant representation by contrastive learning strategy. On one hand, macroscopic images, which share similar shape characteristics in morphology, are introduced to contrast for structure enhancement. On the other hand, different appearance transformations, including color distortion and flittering, are utilized to contrast for texture elimination. In the case where only 1\% of microscopic images are labeled, the proposed method reaches an accuracy of 94.90\% in a generalized testing set.


\end{abstract}
\begin{keywords}
parasite recognition, microscopic image, semi-supervised learning, contrastive learning
\end{keywords}
\section{INTRODUCTION}

Apicomplexans comprise a group of intracellular protozoan parasites, including Plasmodium, Babesia, and Toxoplasma, which are amongst the most prevalent and morbidity-causing pathogens in humans and animals worldwide, and they live inside host cells Erythrocytes. Plasmodium is the causative agent of malaria, which impacts over 200 million individuals and kills over 300,000 children annually. Babesiosis caused by Babesia is a disease with many clinical features that are similar to those of malaria. Toxoplasma is estimated to infect one-third of the world's population, which is the leading cause of infectious retinitis in children and is life-threatening in pregnancy and to the immunocompromised \cite{davila2019overview}.

 The risk of illness and death could be significantly reduced with accurate and affordable diagnostic testing. Measuring parasite infection by direct microscopy observation remains relatively widespread as a point-of-care diagnostic in clinical and epidemiological settings \cite{wu2015Comparison}.

Recently, deep learning techniques become a popular choice in both computer vision and the medical imaging community \cite{Mahmud2018Applications}. For parasites and Erythrocytes recognition, deep learning models have obtained impressive results. A deep learning model using hand-craft features in the shape to classify healthy and abnormal Erythrocytes \cite{lee2014cell}. Well-known Convolutional Neural Networks (CNNs), including Inception v3 \cite{penas2017malaria}, LeNet, AlexNet and GoogLeNet \cite{dong2017evaluations}, are used to identify Plasmodium parasites. Furthermore, a solution for multiple apicomplexan parasites and Erythrocytes \cite{li2020parasitologist} is given based on the fact that Toxoplasma, Babesia, Plasmodium, and Erythrocyte are variant in morphology under microscopy.

However, the outlined methods have been designed in the prerequisite that all the training data has accurate human annotations. In practical clinical scenarios, lack of labels and annotations is a common and fatal problem \cite{kohli2017medical} due to the shortage of specialists in diagnostic imaging despite the increasing spread of equipment.

In addition, there are two challenges for microscopic image recognition. The first is that salient structures of microscopic images are more fuzzy and intricate than real-world objects', which leads to that microscopic images are not as distinguishable as real-world images. The second is that insignificant textures, like image background staining, lightness, or contrast level, are variant a lot in samples from different clinical scenarios, while the available training data \cite{DCTLmicro} is of a similar pattern in each category, which may lead to poor generalization in the real-world application.

In this paper, we propose a semi-supervised learning (SSL) solution for the classification of apicomplexan parasites and host cell microscopic images, considering that a small number of reliable annotations are relatively affordable and it is easy to acquire a large amount of unlabeled data. The proposed method contains data pre-processing, a CNN-based feature extractor, and a multilayer perceptron (MLP) classifier.


To address these challenges,  we learn a distinguishable and appearance-invariant representation by using the contrastive learning strategy.
For structure enhancement, the macroscopic images which share similar shape characteristics in morphology are introduced for contrast. We encourage the similarity between the microscopic images and their corresponding macroscopic images, and the dissimilarity between the unrelated images. For texture elimination, different appearance transformations, including color distortion and flittering, are utilized for contrast. We maximize the similarity between two different appearance transformed views of the same image, while simultaneously minimizing the similarity between different images.


\section{Problem Statement}

Microscopic images from parasite infection samples of preserved slides stained with Giemsa \cite{DCTLmicro} including 5758 Plasmodium, 5741 Toxoplasma, 5878 Babesia, and 6981 Erythrocytes, which are non-infected host cells,  are used as Micro data. 1000 Micro data in each category are split as a testing set. To evaluate the generalization ability of a model, the testing set is transformed to a color dropped version, which will never be used as data augmentation in the training phase. For the training phase, we suppose that only 50 Micro data in each category are labeled, about 1\% of the whole data set, in a practical clinical scenario. 

According to the parasitologist-level knowledge, Plasmodium is ring-shaped, Toxoplasma is generally banana-shaped, Babesia is typically double-pear-shaped and  Erythrocytes resemble an apple. We photo 500 macroscopic images each of ring, banana, double-pear, and apple as Macro data, which best match the Micro objects of interest, instead of using images from the Internet \cite{li2020parasitologist}, since there are usually irrelative items and confusing background, and collecting from the Internet takes much more effort than photoing. Macro data share the same label with their corresponding Micro data.

\begin{figure}[h]
\centering\includegraphics[width=8cm]{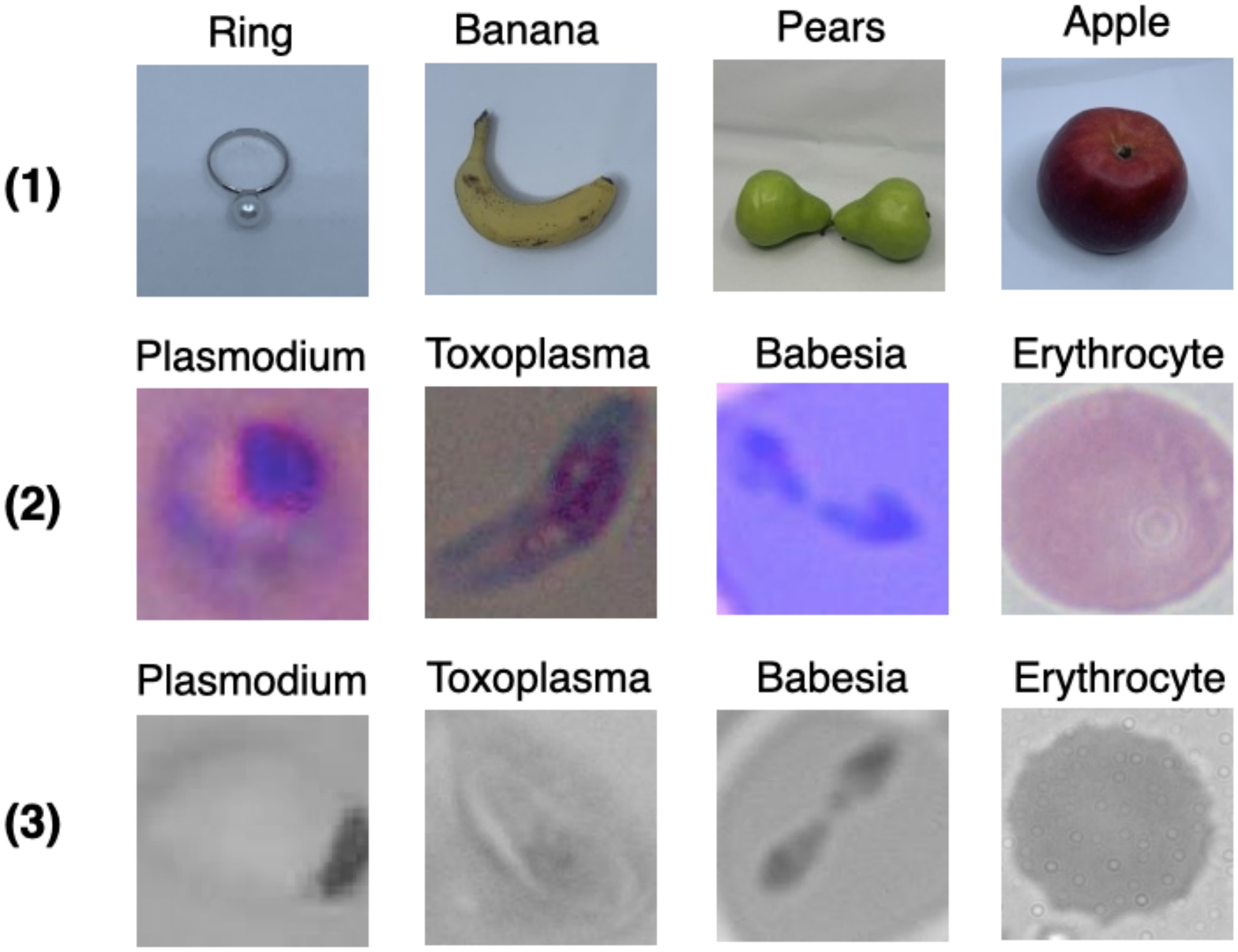}
\caption{Samples from the training set and testing set; (1) Macro data for training, which is full labeled. (2) Micro data for training, which is weakly labeled. (3) Testing data, which is color dropped Micro data.}
\label{data}
 \end{figure}

 Therefore, we have 2000 Macro data $\mathcal{X}_S=\{(x_s,y_s)\}_{s=1}^S$ which is fully labeled with $y_s\in \{0,1,2,3\}$ as source data, 200 labeled Micro data $\mathcal{X}_T=\{(x_t,y_t)\}_{t=1}^T$ with $y_t\in \{0,1,2,3\}$ as target data, and 20158 unlabeled Micro data $X_U = \{x_u\}_{u=1}^U$  in the training set. Fig.\,\ref{data} shows some samples.

\begin{figure}[h]
\centering\includegraphics[width=8.5cm]{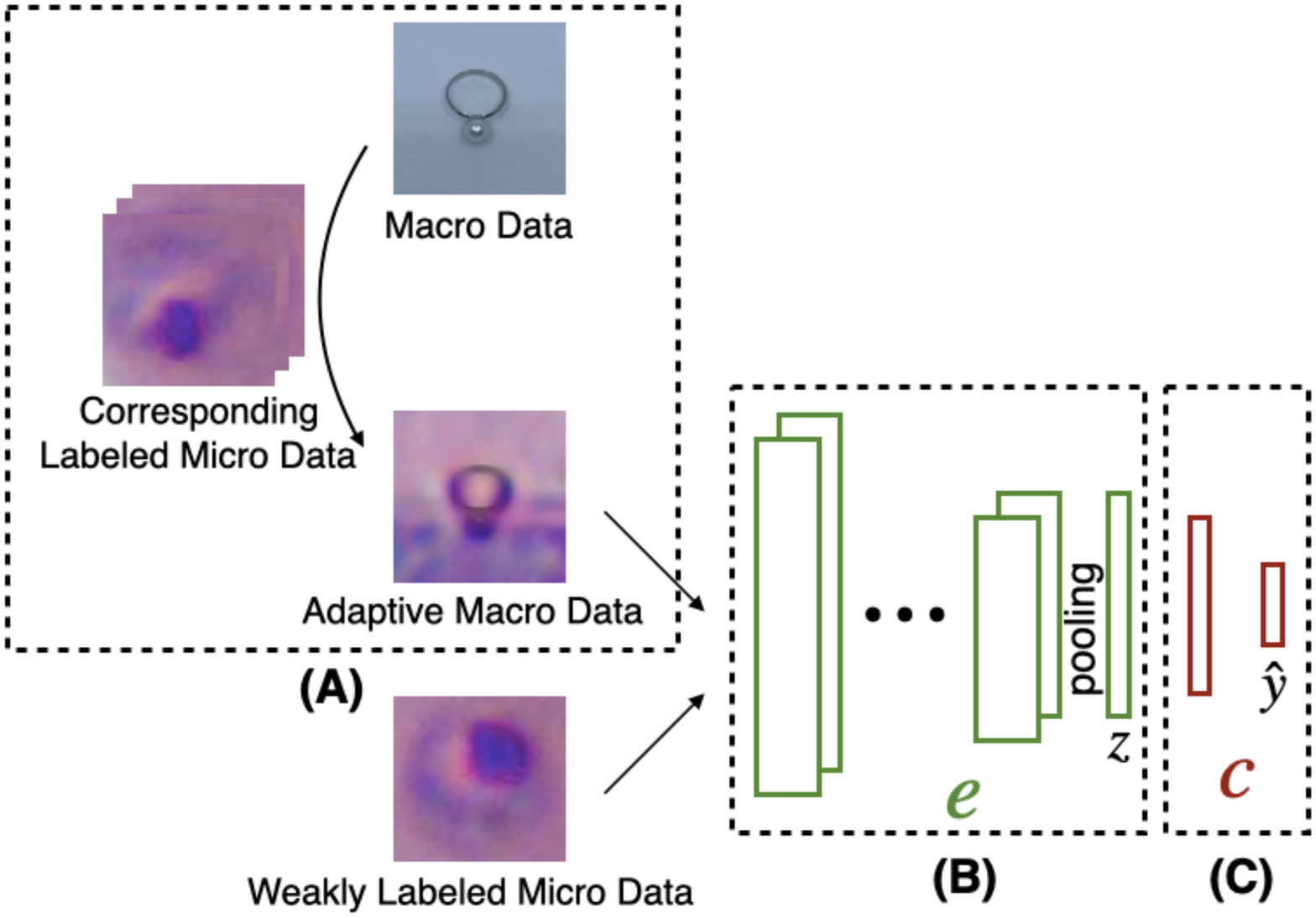}
\caption{Framework of proposed method; (A) Data pre-processing for Macro data. (B) A feature extractor $e$ trained by both labeled and unlabeled data. (C) A  MLP classifier with a ReLU hidden layer $c$  trained by labeled data.}
\label{framework}
 \end{figure}
 
As illustrated in Fig.\,\ref{framework}, here are three modules in our proposed method. Firstly, Macro data is transformed to its adaptive version with the style of corresponding Micro data, as data pre-processing. Then a feature extractor is trained by all the training sets, where the contrastive learning strategy is employed to enhance the salient structure and eliminate the insignificant texture. Finally,  with the learned feature extractor fixed, an MLP classifier is trained by the labeled data including adaptive  Macro data and labeled Micro data.




\section{Methodology}

\subsection{Data Pre-processing}

\begin{figure*}[htbp]
\centering\includegraphics[width=14cm]{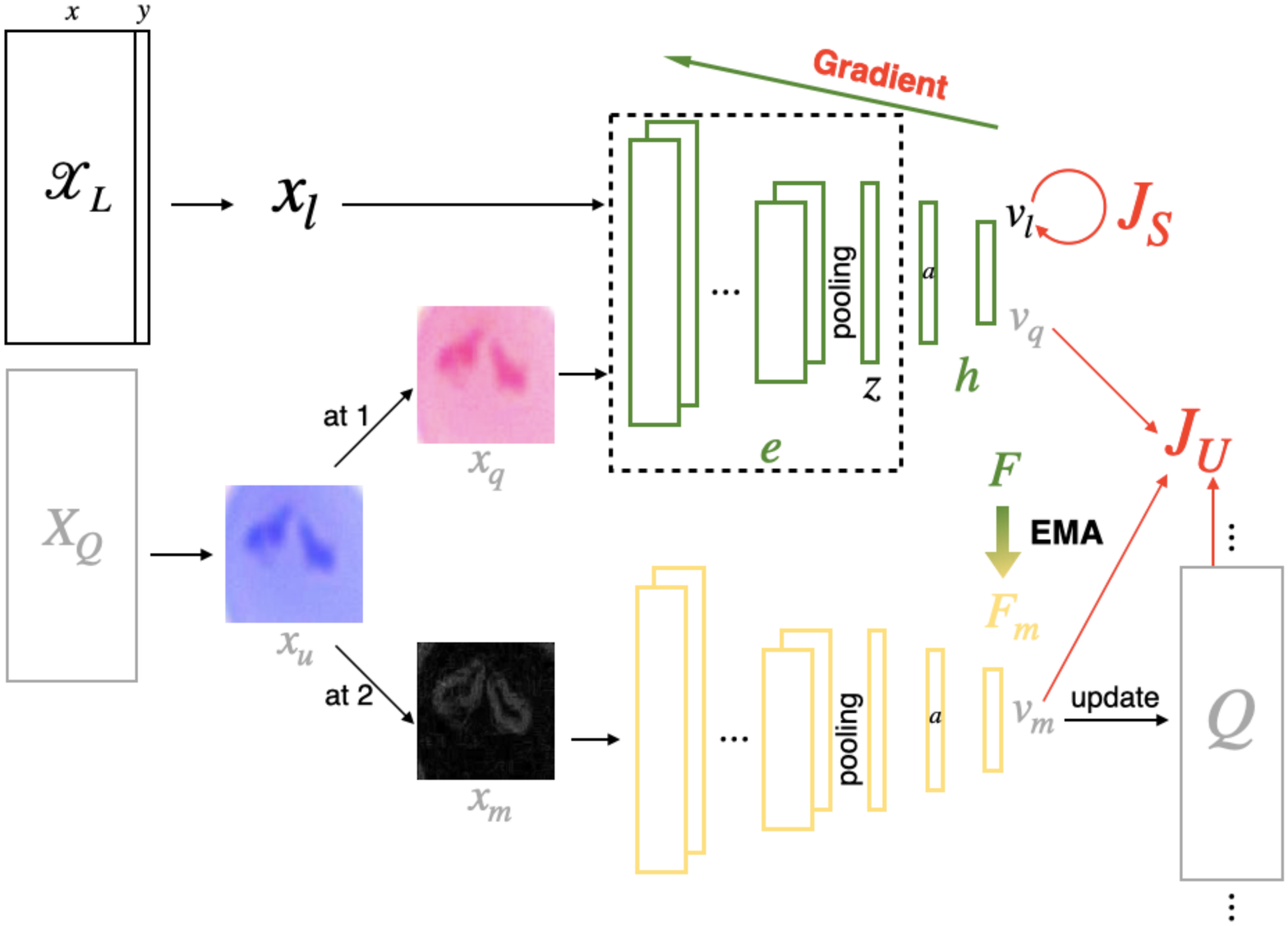}
\caption{Train the feature extractor $e$ using  a contrastive learning strategy. The network $F$, including the feature extractor $e$ and a nonlinear projection head $h$, is updated by the back-propagation on a combination of supervised loss $J_S$ and unsupervised loss $J_U$. An offline network $F_m$ is used as a memory trick, which is an exponential moving average (EMA) of $F$, with a discrete dictionary $Q$ on metric embeddings $v_m$ maintained as a queue. $J_S$ is defined by aligning the labeled Micro data with corresponding Macro data. $J_U$ is defined by aligning two views of different appearance transformation(at) of the same image, where at1 is random color distortion and at2 is Sobel flittering for example.
For the learned representation $z$, $J_S$ contributes to the distinguishability and $J_U$ contributes to the appearance-invariance.}
\label{re_al} 
 \end{figure*}

Real-world objects have clearer shapes and are easier to discriminate. However, if simply use full labeled Macro data and weakly labeled Micro data together in the training phase, the model may perform worse due to the different data distributions, known as ``domain shift" \cite{yao2015semi}.

 Since the shape of Macro data is clearer, while Micro data accounting for a large proportion in quantity. We try to adapt Macro data to appear as if drawn from the Micro domain to realize the visual alignment, as shown in Fig.\,\ref{framework}(A). 
 
 In this subsection, we take rendering the ring image with Plasmodium style as an example. The low-level features of overall the labeled Plasmodium data should be separated as Plasmodium style, and the high-level content in terms of the object in the ring image should be extracted and recombined with Plasmodium style to produce an adaptive ring image. 


Firstly a pre-trained CNN is used to extract and store the features. Content features are  from the ring image $(x_s, 0)$, and style features are from all the labeled Plasmodium images $\mathcal{X}_t = \{(x_t,0)\}_{t=1}^p$. The features on CNN layer $l$  could be stored in a matrix $F^l\in \mathbb{R}^{N_l\times H_l \times W_l }$, where $N_l$ is the number of distinct filters, $H_l$ and $W_l$ denotes the height and width of the feature map respectively. The style  on CNN layer $l$ could be represented by Gram matrix $G^l \in  \mathbb{R}^{N_l\times N_l}$, where $G^{l,ij} = F^{l,i} \odot F^{l,j}$ is the inner product between the vectorized feature maps  $i$ and $j$ in layer $l$ \cite{gatys2015texture}.  

Secondly, the overall loss function 
   \begin{equation}
  \mathcal{L}_{\textit{pre}} = J_{content}(x_a, x_s)  + \lambda_s \cdot J_{style}(x_a, \mathcal{X}_t)
  \label{val}
   \end{equation}
is minimized to update the image $x_a$ iteratively until it simultaneously matches the content features of $x_s$ and the style features of $\mathcal{X}_t$, where $x_a$ can be initiated as a white noise image and  $\lambda_s$ is a coefficient. Since our goal is structure enhancement, which requires preserving the semantic content precisely, we empirically set a small weight of $ \lambda_s = 10^{-3}$ for this purpose.

 
 Based on the fact that deeper convolutional layers respond to higher semantics \cite{gatys2016image}, $ J_{content}$ and $ J_{style}$ are defined as 
   \begin{equation}
   \begin{split}
 J_{content}(x_a, x_s) = & \frac{1}{2}\sum_{l\in L_h} w^l\cdot  (F_a^l -F_s^l)^2\\
 J_{style}(x_a, \mathcal{X}_t)  = &\frac{1}{2}\sum_{l\in L_o}w^l\cdot (G_a^l -\bar{G_t^l})^2
 \end{split}
   \end{equation}
where $L_h$ are the higher CNN layers which capture the high-level content in terms of objects, $L_o$ are all the other CNN layers, $w^l$ is the weight of  layer $l$. Specifically, we use $\bar{G_t^l} = \frac{\sum_t G^{l}_t}{p}$  as the average style features in the Micro domain instead of using the style of a single Micro data. 

With the usage of labeled Micro data, all the Macro data  $\mathcal{X}_S=\{(x_s,y_s)_{s=1}^S\}$  is transferred to its adaptive version  $\mathcal{X}_A=\{(x_a,y_s)_{s=1}^S\}$  with corresponding Micro style.


\subsection{Feature Extractor Learned by contrastive learning strategy}


We aim to learn a  distinguishable and  appearance-invariant representation $z$ for the downstream classification task using a small number of  labeled Micro data  $\mathcal{X}_T=\{(x_t,y_t)\}_{t=1}^T$ and a large amount of unlabeled Micro data $X_U = \{x_u\}_{u=1}^U$, with the additional adaptive Macro data  $\mathcal{X}_A=\{(x_a,y_s)_{s=1}^S\}$. We denote all the labeled data as $\mathcal{X}_L = \mathcal{X}_T \cup \mathcal{X}_A$, and all the  data except labels as $X = X_T  \cup X_A \cup X_U$.

Fig.\ref{re_al} illustrates the proposed method to train the feature extractor $e$ in Fig.\,\ref{framework}(B). Our model denoted as $F(x;\Theta): \mathcal{X}\to \mathbb{R}^{|v|}$ can be decomposed further into a CNN based feature extractor $e(x;\theta_e):  \mathcal{X}\to \mathcal{Z}$ and a MLP projection head $h(z;\theta_h):  \mathcal{Z}\to \mathcal{V}$ conceptually \cite{chen2020simple}, where $z$ is the representation for downstream tasks and $v$ is the metric embedding for contrastive loss, with respect to an input $x$, and $a(\cdot)$ is a non-linear activation function using $k$-sparse strategy \cite{Makhzani2014kSparseA}  
 \begin{equation}
       a(w_j^Tz)
            = \left\{
             \begin{array}{lr}
             w_j^Tz, & j \in \Gamma = supp_k\{W^Tz\}\\
             0, & j \notin \Gamma = supp_k\{W^Tz\}
             \end{array}
\right. 
\end{equation}
$\Gamma = supp_k\{W^Tz\}$ containing hidden units with top-$k\%$ activation values. 
In addition to the online network $F$, a  offline momentum network $F_m(x;\Theta_m)$ is used as a memory trick \cite{he2020momentum}. 

 The update of $\Theta$ is by back-propagation on a combination of  supervised loss $J_S$ and  unsupervised loss $J_U$
\begin{equation}
\mathcal{L}_{e}= J_S+\lambda \cdot J_U
\end{equation}
while $\Theta_m$ is an exponential moving average(EMA) of $\Theta$
\begin{equation}
\Theta_m\gets \alpha \cdot \Theta_m + (1-\alpha)\cdot \Theta 
\end{equation}
where we follow the empirical experience and set $\alpha = 0.999$.

Considering the problem of similarity matching as a form of dictionary look-up, with similarity measured by the dot product of the metric embeddings, InfoNCE \cite{oord2018representation} is adopted as the form of contrastive loss function due to its efficiency and simplicity, with a well-grounded motivation from information theory. Given query $v$, the InfoNCE is the negative log-likelihood
 \begin{equation}
 -\log p_v
\end{equation}
where the likelihood is
 \begin{equation}
p_v =  \frac{\exp(v \cdot v^+/\sigma)}{\sum_i\exp(v\cdot v_i/\sigma)}
\end{equation}
 and $\sigma$ is a temperature parameter that controls the concentration level of the distribution \cite{Geoffrey2015Distilling}. The value of InfoNCE is low when $v$ is similar to its positive key $v^+$ , and dissimilar to all the other keys(considered as negative keys $v^-$ for $v$).  The way we define the positive keys and negative keys for all the training data realizes structure enhancement and texture elimination. 

\subsubsection{Structure Enhancement in $J_S$}


Instead of using a simple cross-entropy loss for labeled data $\mathcal{X}_L$, a contrastive loss is used to enhance the structure information by connecting Micro and Macro data directly. Every query embedding $v_t$ from Micro data, its positive keys are sampled from its corresponding adaptive Macro set, and its negative keys are the embeddings of all the other unrelated adaptive Macro data. Thus, Micro data is encouraged to be similar to its corresponding adaptive Macro data, while dissimilar to unrelated adaptive Macro data simultaneously.  And it is the same way for adaptive Macro data. 

Therefore, the supervised loss for $\mathcal{X}_L$ is 
\begin{equation}
J_S =  - \sum_{t\in \Omega_T} \log p_t -  \sum_{a\in \Omega_A} \log p_a
\end{equation}
and we define
 \begin{equation}
 \begin{split}
 p_t & =\sum_{k^+\in \Omega_a }\frac{\exp(v_t\cdot v_{k^+}/\sigma)}{\exp(v_t\cdot v_{k^+}/ \sigma) + \sum_{{k^-} \in \mathbf{C}_{\Omega_A}^{\Omega_a}}\exp(v_t\cdot v_{k^-}/\sigma)}\\
  p_a & =\sum_{k^+\in \Omega_t}\frac{\exp(v_a\cdot v_{k^+}/\sigma)}{\exp(v_a\cdot v_{k^+}/ \sigma) + \sum_{{k^-} \in \mathbf{C}_{\Omega_T}^{\Omega_t}}\exp(v_a\cdot v_{k^-}/\sigma)}
 \end{split}
 \end{equation}
 where $t$ is the index of Micro data and $\Omega_a$ are all the indexes of its corresponding adaptive Macro data,  $a$ is the index of adaptive Macro data and $\Omega_t$ are all the indexes of its corresponding Micro data. $\Omega_T$ are all the indexes of Micro data and $\Omega_A$ are all the indexes  of adaptive Macro data.

\subsubsection{Texture Elimination in $J_U$}
To eliminate the insignificant texture, two kinds of appearance transformations are introduced, color distortion (which randomly changes the hue, lightness, and saturation of an image) and flittering (such as Sobel, Scharr, Laplacian).   For every $x_u$ in $X$, it is randomly transformed to two different views $x_q, x_m$, and 
we define the positive pair $p^+(v_q,v_m)$,  naturally, all the other images are as the negative keys for $x_u$. 

Since InfoNCE benefits from more negative keys, we apply an offline momentum network $F_m(x;\Theta_m)$ and maintain a discrete dictionary $Q$ on metric embeddings $v_m$ as a queue \cite{he2020momentum}. The metric embeddings of the current mini-batch are enqueued, and the oldest are dequeued.  Introducing $Q$ decouples the dictionary size from the mini-batch size, which can be much larger than the mini-batch size. $\Theta_m$ is the exponential moving average copy from online network $\Theta$ which smooths the learning dynamics.
The unsupervised loss for $X$ is 
\begin{equation}
 J_U = -\sum_u \log p_u
\end{equation}
and we define
\begin{equation}
p_u = \frac{\exp (v_q\cdot v_m/\sigma)}{\exp (v_q\cdot v_m/\sigma) + \sum_i \exp (v_q\cdot Q_i/\sigma )}
\end{equation}
where $v_q = F(x_q;\Theta)$, $v_m = F_m(x_m;\Theta_m)$, and $i$ is the index of the metric embeddings in the maintained  queue under the current state.

\subsection{MLP Classifier}
 
 All the labeled data $\mathcal{X}_L$ is used to train a non-linear MLP classifier $c$ in Fig.\,\ref{framework}(C), which has a ReLU hidden layer, with the learned feature extractor $\hat{e}$ fixed.
 
\section{Experiments}

\subsection{Evaluation Metrics}
We use accuracy (\textit{AC}), F1-measure (\textit{F1}) of sensitivity (\textit{SE}) and  specificity (\textit{SP}), and Jaccard index (JA)  to evaluate the performance
\begin{equation}
    \begin{split}
 AC &= \frac{TP+TN}{TP+FP+TN+FN}\\
    F1 &= \frac{2 \times SE \times SP}{SE + SP}\\
        JA &= \frac{TP}{TP + FP + FN}
    \end{split}
\end{equation}
where $ SE =\frac {TP}{TP+FN}$, $SP= \frac{TN}{TN+FP}$, $TP$, $FP$, $TN$, $FN$ are true positive, false positive, true negative, and false negative, respectively. 

\subsection{Implementation}
We adopt DenseNet-121 \cite{huang2017densely} architecture as the  CNN backbone. In particular, we only include the five convolutional layers in the set, i.e., $L = \{conv1, dense1c, dense2d, dense3f, dense4c\}$ for rendering the Macro data with Micro style, as the representations of these layers, in general, have the highest capability in each scale.  

For a fair comparison, DenseNet-121 architecture is employed as the CNN backbone for all the methods, whose last fully-connected layer (after global average pooling) has a fixed-dimensional output (1024-D). The MLP projection head $h$ leads to 128-D $v$ with a 256-D hidden layer. The ReLU hidden layer of MLP classifier $c$ is 256-D. In particular, we set temperature $\sigma=0.08$ and $k$-sparse 20\%. The batch size is 256, and the Queue size is 4096. The network was trained by SGD algorithm with a learning rate of 0.0001 and a momentum of 0.9. All the experiments are trained for 500 iterations and three repeat training are performed. 


\subsection{Result of Proposed Method}

In this part, we report the performance of the supervised baseline and our method trained with only 50 randomly selected  Macro data in each category, 200 labeled Micro data $\mathcal{X}_T$, and 20158 unlabeled Micro data $X_U$ in Table.\,\ref{basic_results}.  

\begin{table}[h]
\caption{Comparison of supervised learning and semi-supervised learning.  `l' denotes labeled data, `u' denotes unlabeled data.}
\label{basic_results}
\begin{center}
\begin{tabular}{c|c|c|c|c|c}
\hline 
\multicolumn{2}{c|}{Training data} & \multirow{3}{*}{Method} & \multicolumn{3}{c}{Evaluation(\%)}\tabularnewline
\cline{1-2} \cline{2-2} \cline{4-6} \cline{5-6} \cline{6-6} 
Macro & Micro &  & \multirow{2}{*}{AC} & \multirow{2}{*}{F1} & \multirow{2}{*}{JA}\tabularnewline
\cline{1-2} \cline{2-2} 
L & L/U &  &  &  & \tabularnewline
\hline 
\hline 

50{*}4 & 50{*}4/0 & super & 88.12 & 91.51 & 78.64\tabularnewline
\hline 
50{*}4 & 50{*}4/20158 & ours & \textbf{94.90} & \textbf{96.52} & \textbf{90.40}\tabularnewline
\hline 
\end{tabular}
\end{center}
\end{table}

\begin{figure}[h]
\centering\includegraphics[width=7.5cm]{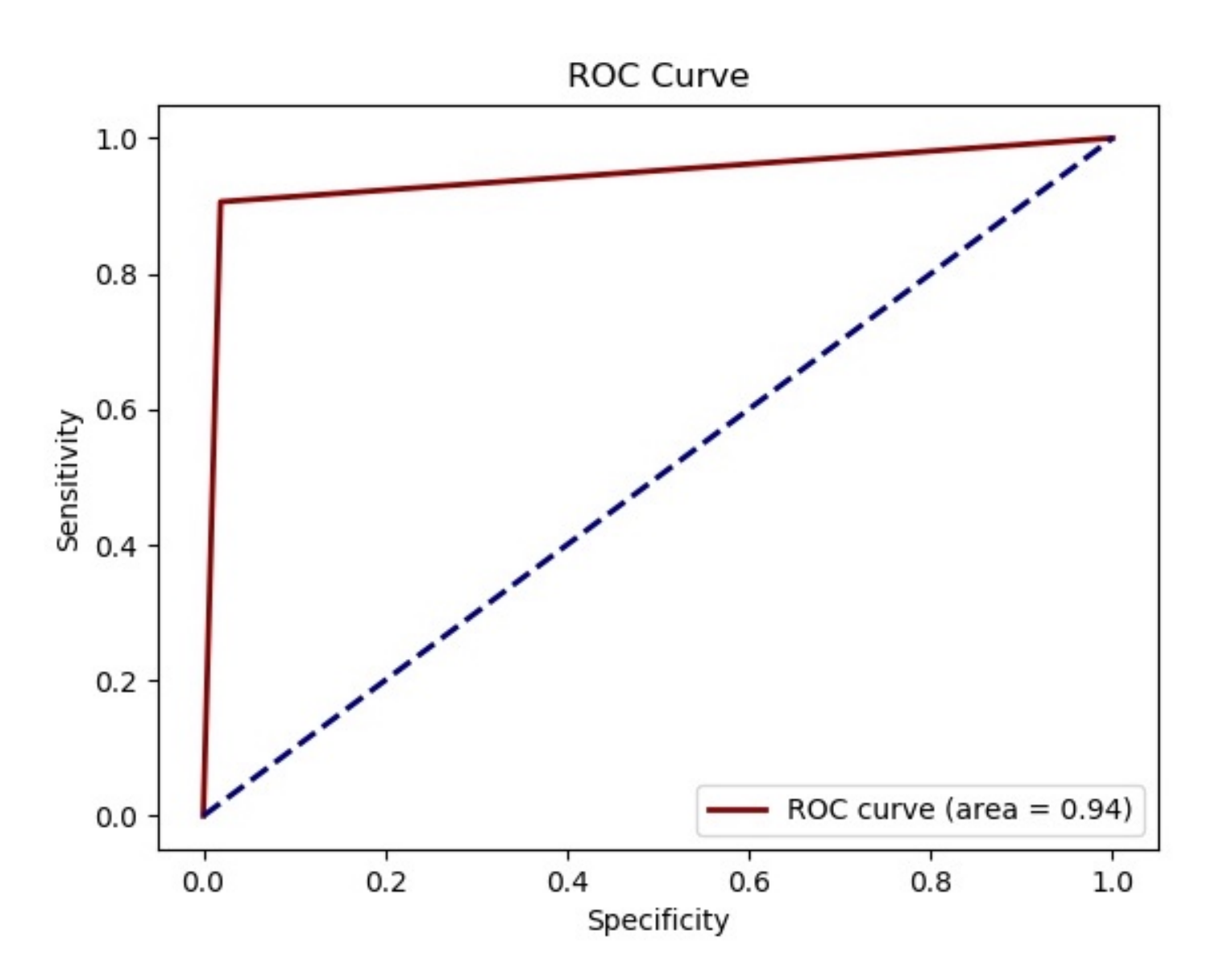}
\caption{Overall ROC result of the proposed method. }
\label{roc_result} 
 \end{figure}

It is obvious that our semi-supervised method achieves higher performance on all the evaluation metrics, with 6.78\%, 5.01\%, and 11.76\% improvements on AC, F1, and JA, respectively, compared with the supervised-only method. Our method,  trained with 1\% labeled Micro data and 10\% Macro data, reaches an accuracy of 94.90\% under a generalized testing set. It is comparable to the reported 95.7\% (AC) in  \cite{li2020parasitologist}, which is trained in a full supervised scenario.

As illustrated in Fig.\,\ref{roc_result}, a receiver operating characteristic (ROC) curve is created to visualize the classification performance of our proposed method for all four classes, and an overall area under the curve (AUC) value is computed to summarize the diagnostic performance.   The proposed model achieves an overall AUC value  of 0.94.

\subsection{Effectiveness of Macro Data Introducing}

We analyze how introducing Macro data helps the final performance, and the results are listed in Table.\,\ref{ablationoF1ac}.

\begin{table}[h]
\caption{Ablation of Macro data Introducing. `l' denotes labeled data, `u' denotes unlabeled data, `o' denotes original version, `a' denotes adaptive version.}
\label{ablationoF1ac}
\begin{center}

\begin{tabular}{c|c|c|c|c|c}
\hline 
\multicolumn{2}{c|}{Training data} & \multirow{3}{*}{Method} & \multicolumn{3}{c}{Evaluation(\%)}\tabularnewline
\cline{1-2} \cline{2-2} \cline{4-6} \cline{5-6} \cline{6-6} 
Marco & Micro &  & \multirow{2}{*}{AC} & \multirow{2}{*}{F1} & \multirow{2}{*}{JA}\tabularnewline
\cline{1-2} \cline{2-2} 
L & L/U &  &  &  & \tabularnewline
\hline 
\hline 
0 & 50{*}4/0 & \multirow{3}{*}{super} & 86.68 & 90.44 & 76.37\tabularnewline
\cline{1-2} \cline{2-2} \cline{4-6} \cline{5-6} \cline{6-6} 
50{*}4(o) & 50{*}4/0 &  & 88.12 & 91.51 & 78.64\tabularnewline
\cline{1-2} \cline{2-2} \cline{4-6} \cline{5-6} \cline{6-6} 
50{*}4(a) & 50{*}4/0 &  & 88.85 & 92.04 & 79.81\tabularnewline
\hline 
0 & 50{*}4/20158 & \multirow{3}{*}{ours} & 92.10 & 94.52 & 85.39\tabularnewline
\cline{1-2} \cline{2-2} \cline{4-6} \cline{5-6} \cline{6-6} 
50{*}4(o) & 50{*}4/20158 &  & 92.97 & 95.16 & 87.04\tabularnewline
\cline{1-2} \cline{2-2} \cline{4-6} \cline{5-6} \cline{6-6} 
50{*}4(a) & 50{*}4/20158 &  & 94.90 & 96.52 & 90.40\tabularnewline
\hline 
\end{tabular}
\end{center}
\end{table}

The first three rows are the results of the supervised method, it helps a lot when introducing Macro data even in its original version. However, the results of the semi-supervision scenario in the last three rows tell a different story. There is a tiny improvement when introducing original Macro data, while an impressive improvement when the Macro data is transformed to its adaptive version.

These results conform to our assumption that `domain shift' is remarkable in a semi-supervision scenario since there are a large amount of unlabeled data being trained together with the labeled data.

\subsection{Comparison With Other Semi-supervised Methods}

As it is much easier to collect real-world objects naturally labeled than annotating microscopic parasites, we analyze the performances with the different numbers of Macro data, where two state-of-the-art semi-supervised learning methods are compared as counterparts.

\begin{figure}[h]
\centering\includegraphics[width=8cm]{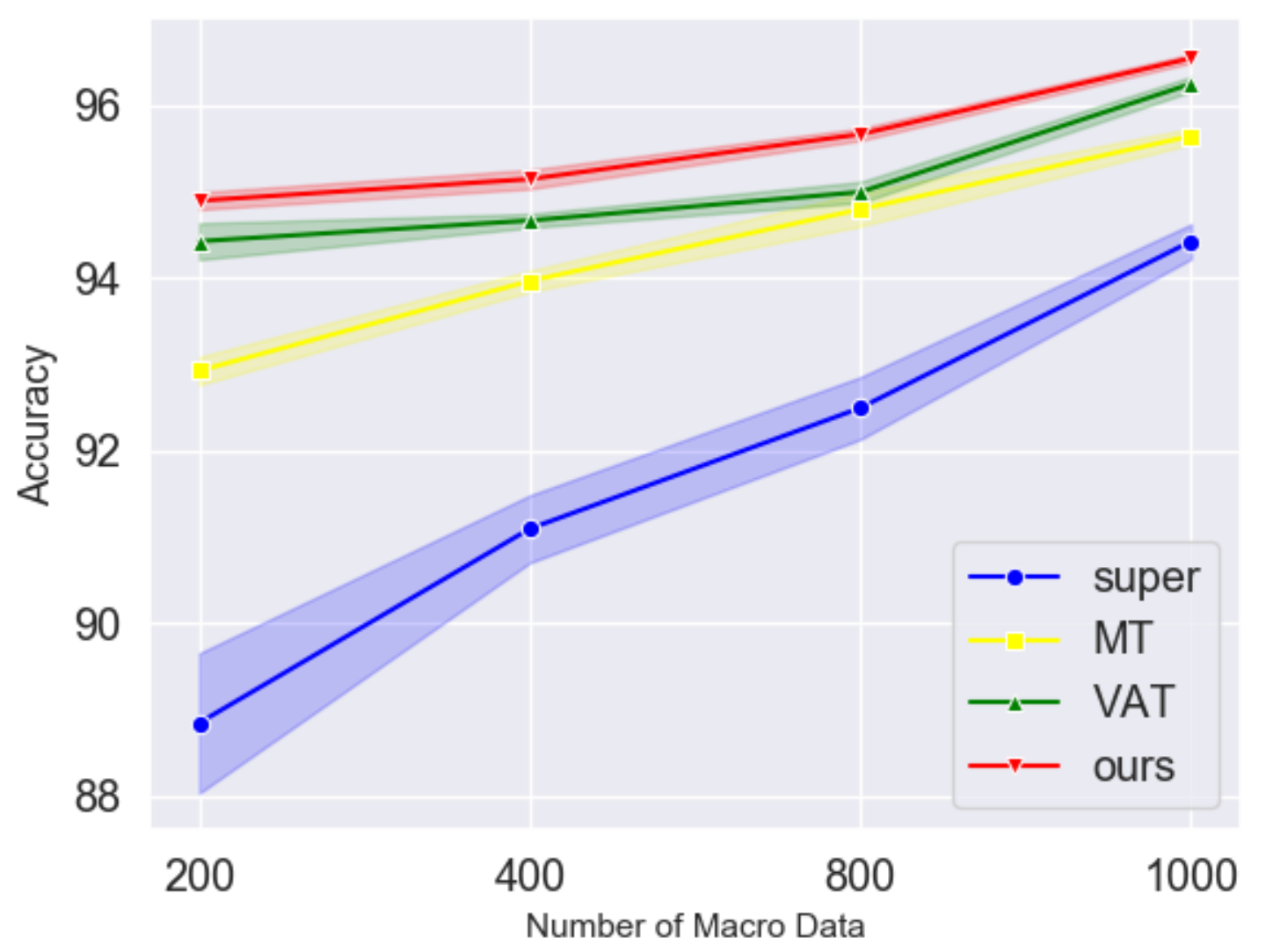}
\caption{Accuracy with different number of Macro data. }
\label{diffl} 
 \end{figure}

Mean Teacher (MT) \cite{Laine2017temporal} used the same EMA strategy as our proposed method from a student network to a teacher network, but simply encourage consistent representations from the same input, without any comparison with others. 
With the same goal of consistency regularization, Virtual Adversarial Training (VAT) \cite{Miyato2018virtual} directly approximated a tiny perturbation to add to input which would most significantly affect the output of the prediction function, instead of using data augmentation. Quantitative results are in Table.\,\ref{ablationofl} and we draw the AC score of the results in Fig.\,\ref{diffl}.

\begin{table}[h]
\caption{Comparison with other SSL methods under different numbers of Marco data. `l' denotes labeled data, `u' denotes unlabeled data.}
\label{ablationofl}
\begin{center}
\begin{tabular}{c|c|c|c|c|c}
\hline 
\multicolumn{2}{c|}{Training data} & \multirow{3}{*}{Method} & \multicolumn{3}{c}{Evaluation(\%)}\tabularnewline
\cline{1-2} \cline{2-2} \cline{4-6} \cline{5-6} \cline{6-6} 
Macro & Micro &  & \multirow{2}{*}{AC} & \multirow{2}{*}{F1} & \multirow{2}{*}{JA}\tabularnewline
\cline{1-2} \cline{2-2} 
L & L/U &  &  &  & \tabularnewline
\hline 
\hline 
\multirow{4}{*}{50{*}4} & \multirow{16}{*}{50{*}4/20158} & super & 88.85 & 92.04 & 79.81\tabularnewline
\cline{3-6} \cline{4-6} \cline{5-6} \cline{6-6} 
 &  & MT & 92.93 & 95.12 & 86.96\tabularnewline
\cline{3-6} \cline{4-6} \cline{5-6} \cline{6-6} 
 &  & VAT & 94.43 & 96.19 & 89.56\tabularnewline
\cline{3-6} \cline{4-6} \cline{5-6} \cline{6-6} 
 &  & ours & \textbf{94.90} & \textbf{96.52} & \textbf{90.40}\tabularnewline
\cline{1-1} \cline{3-6} \cline{4-6} \cline{5-6} \cline{6-6} 
\multirow{4}{*}{100{*}4} &  & super & 91.10 & 93.78 & 83.92\tabularnewline
\cline{3-6} \cline{4-6} \cline{5-6} \cline{6-6} 
 &  & MT & 93.97 & 95.86 & 88.61\tabularnewline
\cline{3-6} \cline{4-6} \cline{5-6} \cline{6-6} 
 &  & VAT & 94.67 & 96.33 & 89.83\tabularnewline
\cline{3-6} \cline{4-6} \cline{5-6} \cline{6-6} 
 &  & ours & \textbf{95.15} & \textbf{96.69} & \textbf{90.72}\tabularnewline
\cline{1-1} \cline{3-6} \cline{4-6} \cline{5-6} \cline{6-6} 
\multirow{4}{*}{200{*}4} &  & super & 92.50 & 94.79 & 86.04\tabularnewline
\cline{3-6} \cline{4-6} \cline{5-6} \cline{6-6} 
 &  & MT & 94.80 & 96.39 & 90.05\tabularnewline
\cline{3-6} \cline{4-6} \cline{5-6} \cline{6-6} 
 &  & VAT & 95.00 & 96.53 & 90.24\tabularnewline
\cline{3-6} \cline{4-6} \cline{5-6} \cline{6-6} 
 &  & ours & \textbf{95.67} & \textbf{97.02} & \textbf{91.67}\tabularnewline
\cline{1-1} \cline{3-6} \cline{4-6} \cline{5-6} \cline{6-6} 
\multirow{4}{*}{500{*}4} &  & super & 94.43 & 96.16 & 89.40\tabularnewline
\cline{3-6} \cline{4-6} \cline{5-6} \cline{6-6} 
 &  & MT& 95.65 & 97.00 & 91.63\tabularnewline
\cline{3-6} \cline{4-6} \cline{5-6} \cline{6-6} 
 &  & VAT & 96.25 & 97.42 & 92.75\tabularnewline
\cline{3-6} \cline{4-6} \cline{5-6} \cline{6-6} 
 &  & ours & \textbf{96.55} & \textbf{97.64} & \textbf{93.32}\tabularnewline
\hline 
\end{tabular}
\end{center}
\end{table}

All these three semi-supervised methods perform better than the supervised baseline consistently, demonstrating that all the SSL methods effectively utilize the unlabeled data and bring performance gains.

As expected, the performances of all the methods increase when more labeled training data are available, and the gap between supervised baseline and semi-supervised methods narrows as more labeled training data are available.

Compared with other methods, we achieve the greatest improvement over the supervised baseline, especially when there are only 200 labels. The comparison shows the effectiveness of our proposed method, compared with other semi-supervised methods. And the success is naturally attributed to contrastive learning strategy.


\section{CONCLUSION and  future work}

In this work, we propose a semi-supervised method using a contrastive learning strategy to classify three apicomplexan parasites and non-infected host cell. The learned representation is more distinguishable by the alignment between Micro data and Macro data, while it is appearance-invariant by encouraging the consistent representation between two different appearance transformed views of the same image.    The results verify the effectiveness of the proposed method. At present, the classifier is relatively simple, it worth further exploring how to take full advantage of the learned representation for classification.








\bibliographystyle{IEEEtran}
\bibliography{ref}

\begin{thebibliography}{10}
\providecommand{\url}[1]{#1}
\csname url@samestyle\endcsname
\providecommand{\newblock}{\relax}
\providecommand{\bibinfo}[2]{#2}
\providecommand{\BIBentrySTDinterwordspacing}{\spaceskip=0pt\relax}
\providecommand{\BIBentryALTinterwordstretchfactor}{4}
\providecommand{\BIBentryALTinterwordspacing}{\spaceskip=\fontdimen2\font plus
\BIBentryALTinterwordstretchfactor\fontdimen3\font minus
  \fontdimen4\font\relax}
\providecommand{\BIBforeignlanguage}[2]{{%
\expandafter\ifx\csname l@#1\endcsname\relax
\typeout{** WARNING: IEEEtran.bst: No hyphenation pattern has been}%
\typeout{** loaded for the language `#1'. Using the pattern for}%
\typeout{** the default language instead.}%
\else
\language=\csname l@#1\endcsname
\fi
#2}}
\providecommand{\BIBdecl}{\relax}
\BIBdecl

\bibitem{davila2019overview}
J.~A.~A. Davila and A.~H. De~Los~Rios, ``An overview of peripheral blood
  mononuclear cells as a model for immunological research of toxoplasma gondii
  and other apicomplexan parasites,'' \emph{Frontiers in cellular and infection
  microbiology}, vol.~9, no.~24, pp. 1--10, February 2019.

\bibitem{wu2015Comparison}
L.~Wu, L.~van~den Hoogen, H.~Slater, P.~Walker, A.~Ghani, C.~Drakeley, and
  L.~Okell, ``Comparison of diagnostics for the detection of asymptomatic
  plasmodium falciparum infections to inform control and elimination
  strategies,'' \emph{Nature}, vol. 528, no.~3, pp. 86--93, December 2015.

\bibitem{Mahmud2018Applications}
M.~{Mahmud}, M.~S. {Kaiser}, A.~{Hussain}, and S.~{Vassanelli}, ``Applications
  of deep learning and reinforcement learning to biological data,'' \emph{IEEE
  Transactions on Neural Networks and Learning Systems}, vol.~29, no.~6, pp.
  2063--2079, June 2018.

\bibitem{lee2014cell}
H.~Lee and Y.-P.~P. Chen, ``Cell morphology based classification for red cells
  in blood smear images,'' \emph{Pattern Recognition Letters}, vol.~49, no.~1,
  pp. 155--161, November 2014.

\bibitem{penas2017malaria}
K.~E.~D. Pe{\~n}as, P.~T. Rivera, and P.~C. Naval, ``Malaria parasite detection
  and species identification on thin blood smears using a convolutional neural
  network,'' in \emph{Proc. of 2017 IEEE/ACM International Conference on
  Connected Health: Applications, Systems and Engineering Technologies
  (CHASE'2017)(Philadelphia)}, July 2017.

\bibitem{dong2017evaluations}
Y.~Dong, Z.~Jiang, H.~Shen, W.~D. Pan, L.~A. Williams, V.~V. Reddy, W.~H.
  Benjamin, and A.~W. Bryan, ``Evaluations of deep convolutional neural
  networks for automatic identification of malaria infected cells,'' in
  \emph{Proc. of 2017 IEEE EMBS International Conference on Biomedical \&
  Health Informatics (BHI'2017)(Orland)}, February 2017.

\bibitem{li2020parasitologist}
L.~Sen, Y.~Qi, J.~Hao, C.-V.~J. A, and Z.~Yang, ``Parasitologist-level
  classification of apicomplexan parasites and host cell with deep cycle
  transfer learning (dctl),'' \emph{Bioinformatics}, vol.~36, no.~16, pp.
  4498--4505, August 2020.

\bibitem{kohli2017medical}
M.~D. Kohli, R.~M. Summers, and J.~R. Geis, ``Medical image data and datasets
  in the era of machine learning-whitepaper from the 2016 c-mimi meeting
  dataset session,'' \emph{Journal of digital imaging}, vol.~30, no.~4, pp.
  392--399, August 2017.

\bibitem{DCTLmicro}
L.~Sen, ``Microdata,'' Mendeley Data,
  \url{https://data.mendeley.com/datasets/7t3y7j6hh8/draft?a=132247f0-5914-49f9-8857-e26e2f1060d8}.

\bibitem{yao2015semi}
T.~Yao, Y.~Pan, C.-W. Ngo, H.~Li, and T.~Mei, ``Semi-supervised domain
  adaptation with subspace learning for visual recognition,'' in \emph{Proc. of
  2015 IEEE Conference on Computer Vision and Pattern
  Recognition(CVPR'2015)(Boston)}, June 2015.

\bibitem{gatys2015texture}
L.~A. Gatys, A.~S. Ecker, and M.~Bethge, ``Texture synthesis using
  convolutional neural networks,'' in \emph{Proc. of Twenty-ninth Conference on
  Neural Information Processing Systems(NIPS'2015)(Montréal)}, December 2015.

\bibitem{gatys2016image}
L.~a. Gatys, A.~S. Ecker, and M.~Bethge, ``Image style transfer using
  convolutional neural networks,'' in \emph{Proc. of the IEEE conference on
  computer vision and pattern recognition(CVPR'2016)(Las Vegas)}, June 2016.

\bibitem{chen2020simple}
T.~Chen, S.~Kornblith, M.~Norouzi, and G.~Hinton, ``A simple framework for
  contrastive learning of visual representations,'' in \emph{Proc. of 37th
  International Conference on Machine Learning(ICML'2020)(online)}, April 2020.

\bibitem{Makhzani2014kSparseA}
A.~Makhzani and B.~Frey, ``k-sparse autoencoders,'' in \emph{Proc. of 2014
  International Conference on Learning Representations(ICLR'2014)(Banff)},
  April 2014.

\bibitem{he2020momentum}
K.~He, H.~Fan, Y.~Wu, S.~Xie, and R.~Girshick, ``Momentum contrast for
  unsupervised visual representation learning,'' in \emph{Proceedings of 2020
  IEEE/CVF Conference on Computer Vision and Pattern
  Recognition(CVPR'2020)(online)}, June 2020.

\bibitem{oord2018representation}
A.~v.~d. Oord, Y.~Li, and O.~Vinyals, ``Representation learning with
  contrastive predictive coding,'' \emph{arXiv preprint arXiv:1807.03748},
  2018.

\bibitem{Geoffrey2015Distilling}
G.~Hinton, O.~Vinyals, and J.~Dean, ``Distilling the knowledge in a neural
  network,'' in \emph{Proc. of 2015 annual conference on neural information
  processing systems(NIPS'2015)(Montreal) Deep Learning and Representation
  Learning Workshop}, 2015.

\bibitem{huang2017densely}
G.~Huang, Z.~Liu, L.~V.~D. Maaten, and K.~Q. Weinberger, ``Densely connected
  convolutional networks,'' in \emph{Proceedings of 2017 IEEE/CVF Conference on
  Computer Vision and Pattern Recognition(CVPR'2017)(Hawaii)}, July 2017.

\bibitem{Laine2017temporal}
S.~Laine and T.~Aila, ``Temporal ensembling for semi-supervised learning,'' in
  \emph{Proc. of Fifth International Conference on Learning
  Representations(ICLR'2017)(Toulon)}, 2017.

\bibitem{Miyato2018virtual}
T.~Miyato, S.-I. Maeda, S.~Ishii, and M.~Koyama, ``Virtual adversarial
  training: A regularization method for supervised and semi-supervised
  learning,'' \emph{IEEE Transactions on Pattern Analysis and Machine
  Intelligence}, vol.~41, no.~8, pp. 1979 -- 1993, August 2018.

\end{thebibliography}

\end{document}